\documentclass{egpubl}

\Poster                 
\electronicVersion 

\ifpdf \usepackage[pdftex]{graphicx} \pdfcompresslevel=9
\else \usepackage[dvips]{graphicx} \fi
\usepackage[T1]{fontenc} 
\BibtexOrBiblatex
\PrintedOrElectronic

\usepackage{t1enc,dfadobe} 

\usepackage{egweblnk}
\usepackage{cite}

\title[VITON-GAN: Virtual Try-on Image Generator Trained with Adversarial Loss]%
      {VITON-GAN: Virtual Try-on Image Generator\\
      Trained with Adversarial Loss}

\author[Shion Honda]
{\parbox{\textwidth}{\centering Shion Honda$^{1,2}$}\\
{\parbox{\textwidth}{\centering $^1$The University of Tokyo\\$^2$The International Research Center for Neurointelligence}}
}

\begin{document}

\teaser{
\includegraphics[height=5cm]{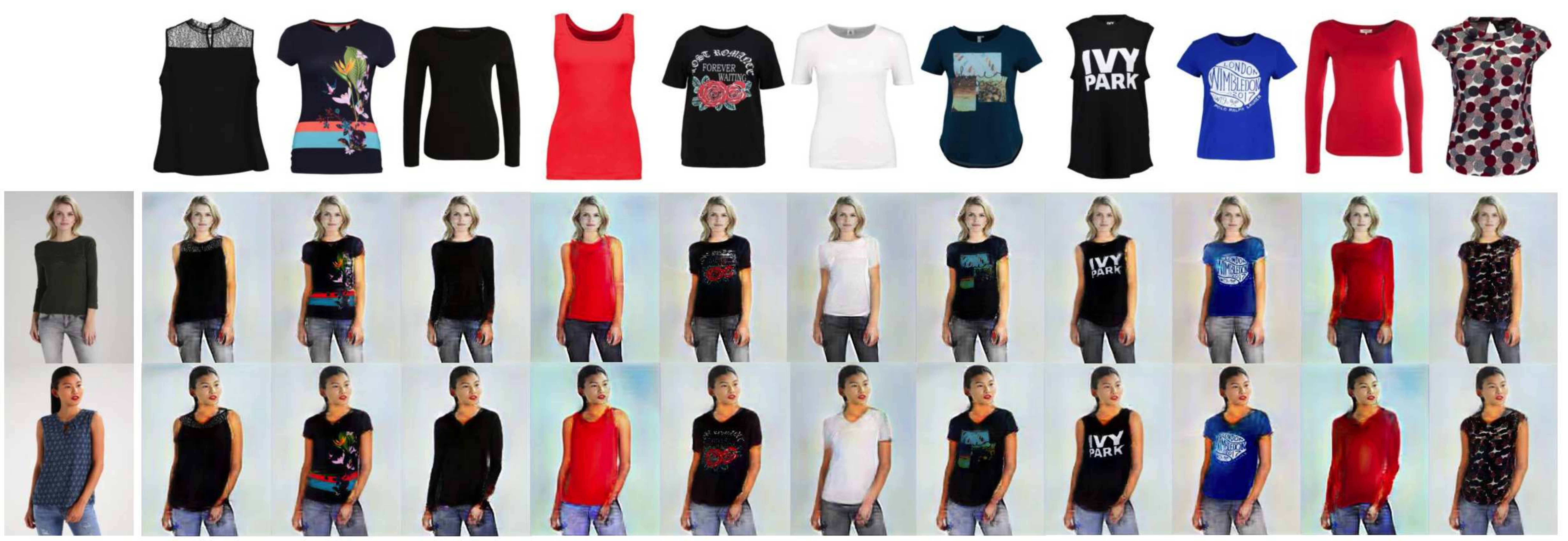}
\centering
\caption{Samples from generated images. The models in the left column virtually wear the clothes from the top row.}
\label{fig:teaser}
}

\maketitle

\begin{abstract}
Generating a virtual try-on image from in-shop clothing images and a model person's snapshot is a challenging task because the human body and clothes have high flexibility in their shapes. In this paper, we develop a Virtual Try-on Generative Adversarial Network (VITON-GAN), that generates virtual try-on images using images of in-shop clothing and a model person. This method enhances the quality of the generated image when occlusion is present in a model person's image (e.g., arms crossed in front of the clothes) by adding an adversarial mechanism in the training pipeline.   

\ccsdesc[300]{Computing methodologies~Image representations}
\ccsdesc[300]{Applied computing~Online shopping}

\printccsdesc   
\end{abstract}  
\section{Introduction}
Despite the recent growth of online apparel shopping, there is a tremendous demand by consumers for buying clothes after trying them on in real shops. If e-commerce sites can offer virtual try-on images from a snapshot of the customer, they can improve their user experience.

To solve this task, previous approaches combined a U-net generator and thin plate spline (TPS) transform \cite{han2018viton}\cite{wang2018toward}. The TPS transform keeps the patterns and letters of the clothes accurate when mapped on the human body. The latest work (CP-VTON) \cite{wang2018toward} used a human parser \cite{Gong_2017_CVPR} and pose estimator \cite{cao2017realtime} in its pipeline to extract the person's representation (explanatory variable) independent of wearing clothes (objective variable). As shown in Figure \ref{fig:sample_good}, however, we report that these methods fail when arms are crossed in front of the clothes (occlusion), generating blurred arms due to reconstruction loss. 

For generating realistic images, generative adversarial networks (GANs) have been successfully used \cite{brock2018large}\cite{Karras_2018}. Unlike other generative models using reconstruction loss such as VAE, GANs are able to generate fine, high-resolution, and realistic images because adversarial loss can incorporate perceptual features that are difficult to define mathematically.

In this paper, we propose an image generator that alleviates the occlusion problem, called Virtual Try-On GAN (VITON-GAN). This generator consists of two modules, the geometry matching module (GMM) and the try-on module (TOM) as was implemented in CP-VTON, except adversarial loss is additionally included in the TOM to address the occlusion problem.
\section{Methods}
\begin{figure}[htb]
  \centering
  \includegraphics[width=.85\linewidth]{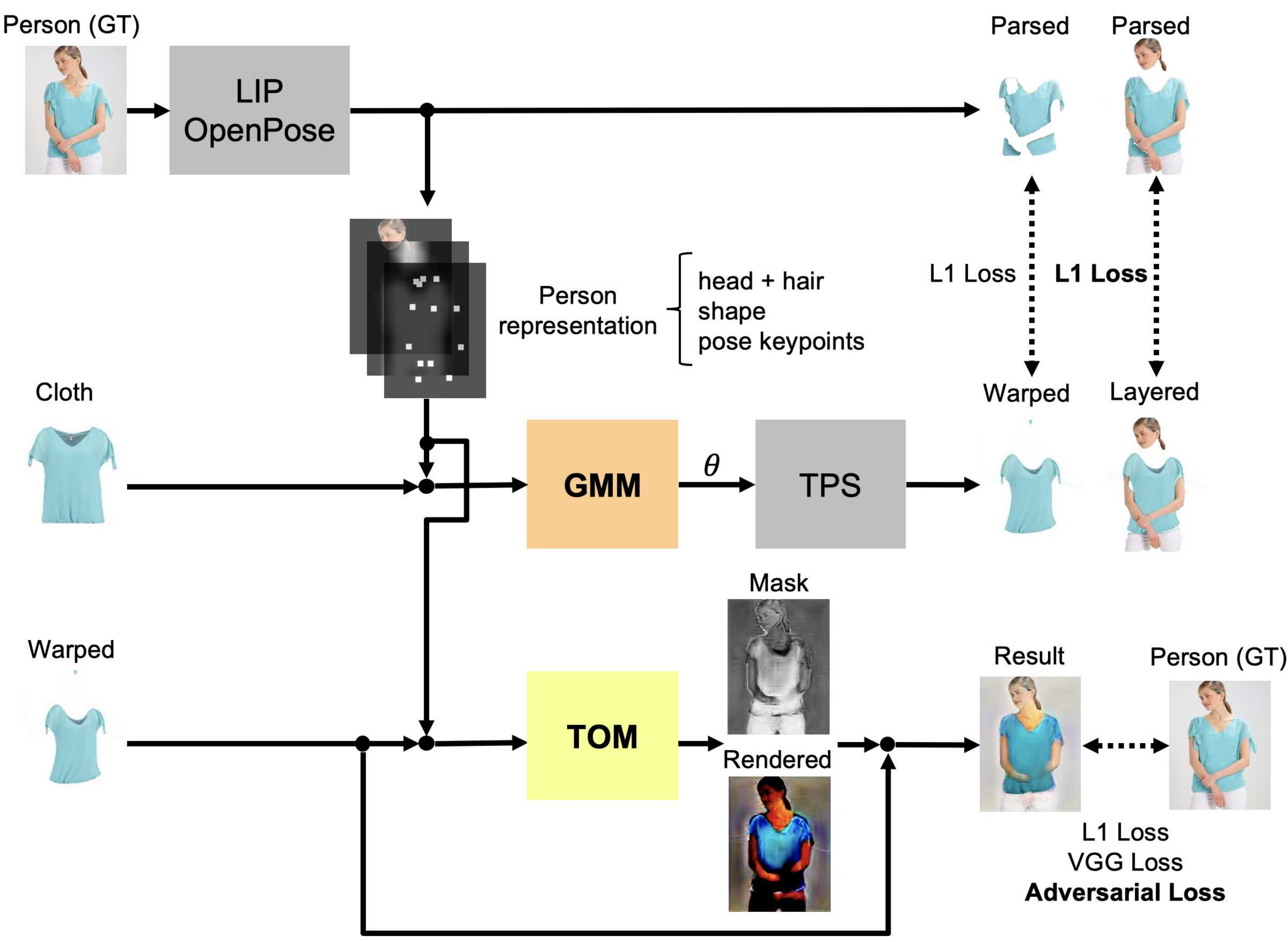}
  \caption{\label{fig:overview}
           Overview of the VITON-GAN training pipeline.}
\end{figure}

The whole training pipeline of VITON-GAN is presented in Figure \ref{fig:overview}. There are three major updates from CP-VTON. First, TOM is trained adversarially against the discriminator that uses the TOM result image, in-shop clothing image, and person representation as inputs and judges whether the result is real or fake. Second, the loss function of GMM includes the L1 distance between the generated and real images of clothes layered on the body. Finally, random horizontal flipping is used for data augmentation. The source codes and the trained model are available at \url{https://github.com/shionhonda/viton-gan}.

\section{Experiments and Results}
To show the effect on the occlusion problem of VITON-GAN, a virtual try-on experiment was conducted using the same dataset as CP-VTON. The dataset contains 16,253 female model's snapshots and top clothing image pairs, which were split into 13,221, 1,000, and 2,032 pairs for training, validation, and test sets, respectively. All result images presented in this paper are from the test set. 

\begin{figure}[htb]
  \centering
  \includegraphics[width=.8\linewidth]{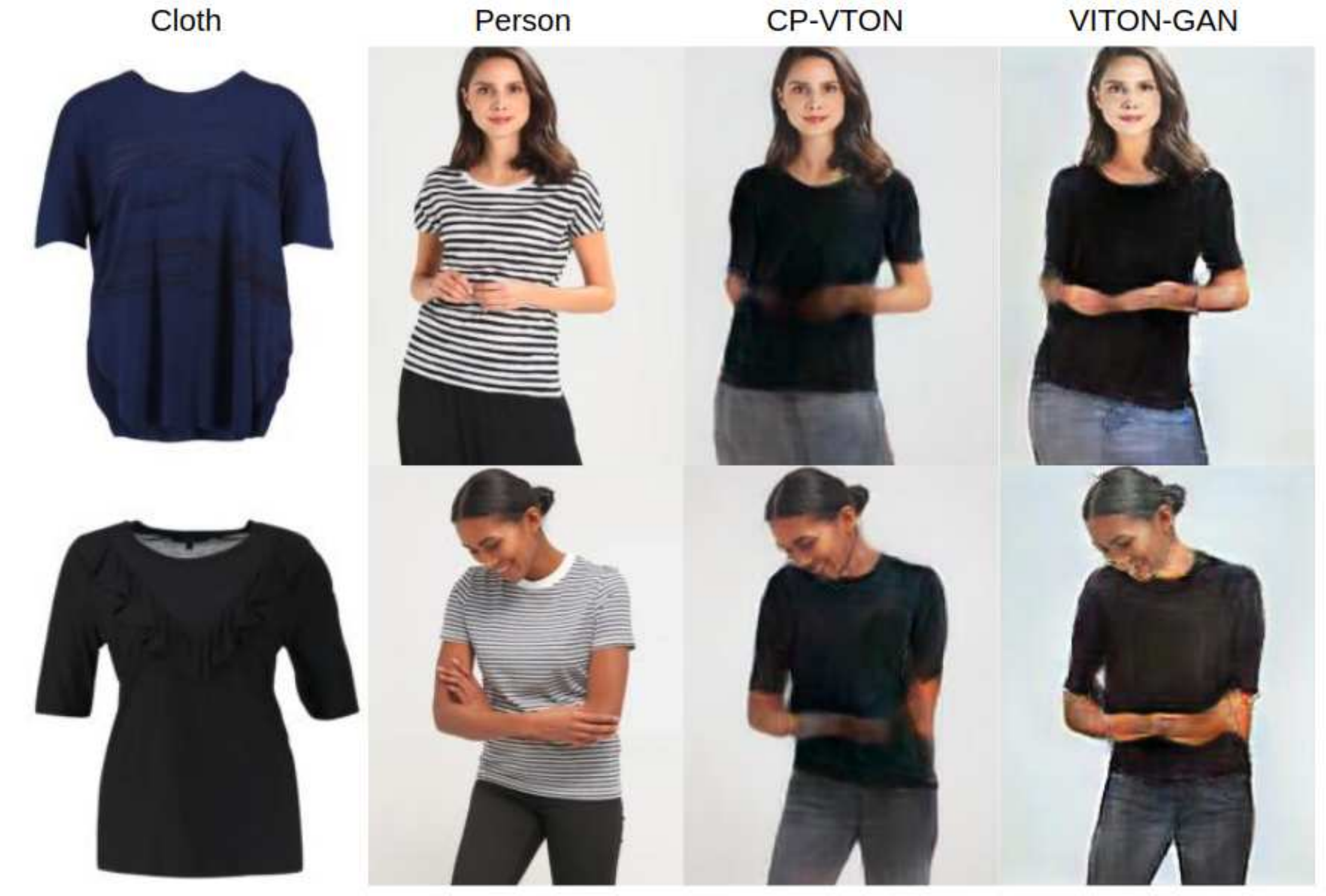}
  \caption{\label{fig:sample_good}
           Successful cases of the proposed method.}
\end{figure}

\begin{figure}[htb]
  \centering
  \includegraphics[width=.8\linewidth]{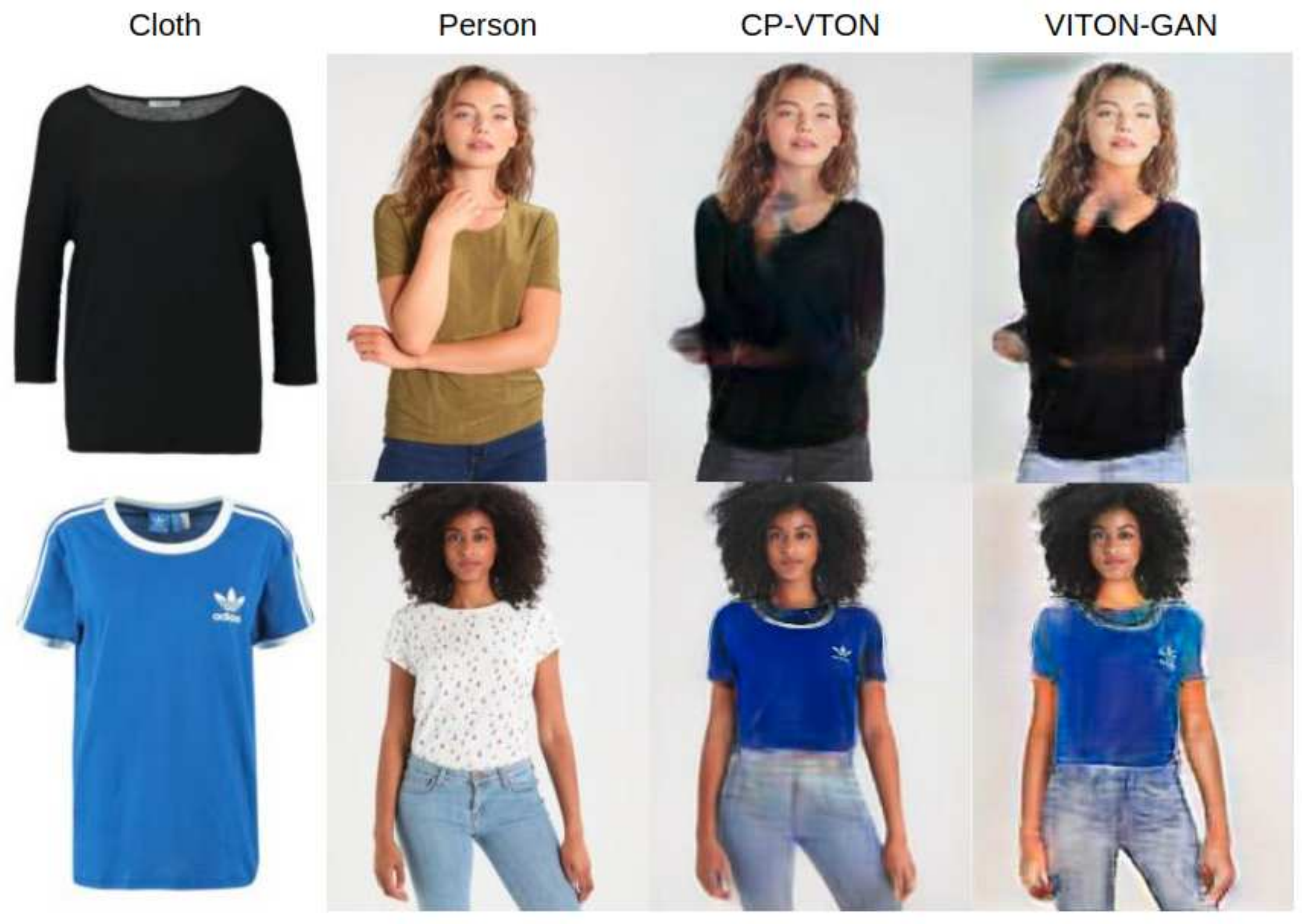}
  \caption{\label{fig:sample_bad}
           Failed cases of the proposed method.}
\end{figure}

As shown in Figure \ref{fig:sample_good}, VITON-GAN generated hands and arms more clearly than CP-VTON in occlusion cases. However, arm generation failed when the model's original clothing was half-sleeve and the tried-on clothing was long-sleeve (see Figure \ref{fig:sample_bad}: upper row). This was because the TPS transform was not able to deal with topological changes that often occurred in the case of occlusion with long-sleeve shirts. Also, although in most cases VITON-GAN generated images as fine as CP-VTON (see Figure \ref{fig:teaser}), it occasionally generated blurred images as shown in the lower row of Figure \ref{fig:sample_bad}.

\section{Conclusions}
Here, we propose a virtual try-on image generator from 2D images of a person and top clothing that alleviates the occlusion problem. Future work will include improving the quality of generated parts of the human body and addressing topological changes in the clothes.

\section*{Acknowledgements}
This work was supported by the Chair for Frontier AI Education, the
University of Tokyo. Also, we would like to thank I. Sato, T. Harada, T. Mano, and C. Yokoyama for correcting this paper.
\bibliographystyle{eg-alpha-doi} 
\bibliography{egbibsample.bib}

\end{document}